\documentclass[11pt]{article}
\usepackage[margin=1in]{geometry}
\usepackage{amsmath,amssymb}
\usepackage{graphicx}
\usepackage{booktabs}
\usepackage{multirow}
\usepackage{hyperref}
\usepackage{authblk}
\usepackage{caption}
\usepackage{xcolor}
\usepackage{natbib}
\usepackage{tabularx}
\usepackage{makecell}
\usepackage{array}
 \usepackage{float}

\title{\textbf{TRUST: Threshold-Recalibrated Uncertainty-Safe Training\\
for Certified Dismissal in Breast Cancer Screening}}
 
\author[1]{Parham Shafie}
\author[1]{Matthew Hamilton}

\author[2]{Edward Kendall}
\author[3]{Gregory Doyle}
\author[1]{Oscar Meruvia Pastor}
\affil[1]{Department of Computer Science, Memorial University of Newfoundland, Canada}
\affil[2]{Division of Biomedical Sciences, Faculty of Medicine, Memorial University of Newfoundland, Canada}
\affil[3]{Cancer Care, Newfoundland and Labrador Health Services, St. John’s, NL, Canada}
\date{}
 
\begin{document}
\maketitle
\begin{abstract}
Reducing the review of clearly cancer-negative screening mammograms could
lower radiologist workload without compromising cancer detection. We propose
a closed-loop threshold-aware training strategy in which the dismissal
threshold is recalculated during training and used to penalize
cancer-positive images that approach the dismissal region. We evaluated the
method on NLBS and RSNA using five controlled training configurations, with
case-level assessment based on a one-sided 99\% Clopper--Pearson upper bound
for cancer prevalence among dismissed cases. The proposed model achieved the
highest case-level dismissal rates at both 98\% and 95\% recall targets. On
NLBS, dismissal reached 19.74\% and 21.70\%, while the cross-entropy baseline
did not meet either recall target. On RSNA, dismissal improved from 7.04\% to
14.31\% and from 13.49\% to 19.69\%. In external RSNA$\to$NLBS evaluation,
the proposed model achieved dismissal rates of 12.95\% and 19.87\% at the
98\% and 95\% recall targets, respectively. These results support closed-loop
threshold-aware training for high-recall selective dismissal.
\end{abstract}

\textbf{Keywords:} safe dismissal, breast cancer screening, selective classification, statistical risk control, mammography

\section{Introduction}

\subsection{Clinical motivation}

In 2024, the U.S. Preventive Services Task Force lowered the recommended starting age for mammography screening from 50 to 40 years and recommended biennial screening for average-risk women aged 40--74~\cite{10.1001/jama.2024.5534}. Screening recommendations vary somewhat across organizations. The American College of Radiology recommends annual mammography beginning at age 40, while the American Cancer Society recommends optional annual screening for women aged 40--44, annual screening for those aged 45--54, and annual or biennial screening from age 55 onward~\cite{MONTICCIOLO20211280,10.1001/jama.2015.12783}.

Every screening mammogram is assigned a Breast Imaging Reporting and Data System (BI-RADS) category~\cite{Destounis2025}. Most examinations are classified as BI-RADS 1 or 2, and these patients return for routine screening. In one screening cohort, 7.5\% of examinations were classified as BI-RADS 0, with higher rates at first screening than at subsequent
screening examinations~\cite{Negrao2025-kr}. This evaluation commonly includes additional mammographic views and targeted breast ultrasound~\cite{Den_Dekker2024-kl}. Based on the additional findings, the examination may then be classified as BI-RADS 1 or 2, BI-RADS 3 requiring short-interval follow-up, or BI-RADS 4 or 5 for which biopsy is recommended~\cite{statpearlsbirads}. Although biopsy provides the definitive diagnosis when required, most recalled findings and a substantial proportion of biopsied findings are ultimately benign~\cite{Hubbard2011-po}. These additional steps increase radiologist workload and imaging use while also contributing to patient anxiety and healthcare costs.

The economic burden of breast cancer screening extends beyond the initial
mammographic examination. A recent Canadian health-system analysis estimated
the cost of a digital screening mammogram at CAD\$86.92 in 2023, including
professional, technical, and administrative expenses~\cite{Wilkinson2025-kp}.
For every 1,000 women followed from age 40, biennial screening between ages
40 and 74 was associated with approximately 16,100 lifetime mammograms,
1,484 imaging recalls, and 191 biopsies, of which approximately 126 were
nonmalignant. This corresponded to about CAD\$1.44 million in screening
costs and an additional CAD\$127,500 in diagnostic costs related to recalled
examinations. More intensive annual screening increased these expenditures,
with approximately CAD\$2.79 million in screening costs and CAD\$241,000 in
recall-related diagnostic costs per 1,000 women. These figures illustrate the
substantial cumulative resource use associated with population-scale
mammography programs.

At the same time, the value of screening cannot be assessed from imaging
costs alone. Earlier cancer detection can reduce treatment expenditure by
shifting diagnosis toward less advanced disease. In the same Canadian
analysis, breast cancer treatment costs ranged from approximately
CAD\$14,500 for stage 0 disease to more than CAD\$500,000 for some stage IV
disease presentations, and increased screening intensity was associated with
lower overall treatment costs~\cite{Wilkinson2025-kp}. In a sensitivity analysis
in which the recall rate was reduced toward the Canadian target of 5\%,
biennial screening from ages 40 to 74 was associated with approximately
CAD\$102,000 in health-system savings per 1,000 women compared with biennial
screening from ages 50 to 74. These findings highlight the importance of
reducing unnecessary downstream screening activity without compromising
cancer detection, providing an economic as well as clinical motivation for
AI-based approaches that can reliably identify examinations unlikely to
require further review.

The growing workload associated with screening mammography has produced an increased interest in using artificial intelligence to reduce the number of examinations that require manual radiologist review. Large retrospective studies have reported deep learning performance comparable to that of individual radiologists across screening populations~\cite{Wu2020-gg, McKinney2020-zk}. More recently, the randomized MASAI trial demonstrated that AI-supported triage can be integrated into clinical screening workflows and reduce radiologist screen-reading workload by 44\% while maintaining the cancer detection rate~\cite{Lang2023-vd}. Follow-up analyses also reported improved screening performance and a non-inferior interval cancer rate compared with standard double reading~\cite{Hernstrom2025-wf, Gommers2026-db}. These findings provide evidence that AI can reduce screening workload while maintaining high cancer detection performance.

However, the safety demonstrated in these studies is based on observed performance within specific populations, workflows, and operating thresholds. This differs from a model-level statistical guarantee that can be independently evaluated, audited, and recalculated when the model is applied to a new setting. Conventional measures such as AUROC, sensitivity, and specificity describe overall discrimination, but they do not directly quantify how many examinations can be removed from manual review while controlling the risk of cancer among those dismissed.

This study therefore focuses on certified dismissal rather than automated diagnosis. The aim is not to replace radiologist judgment or to assume that false-negative cases can be eliminated completely. Instead, we investigate what proportion of screening examinations can be dismissed from manual review while maintaining a finite-sample statistical upper bound on the cancer rate within the dismissed group. Our framework addresses this objective using an exact binomial confidence procedure to quantify the statistical uncertainty in the cancer rate among dismissed cases, as described in Section~\ref{sec:methods}.
Existing selective-classification methods commonly determine an accept/reject
threshold after classifier training or jointly learn prediction and selection
functions~\cite{geifman2017selective,Geifman2019-yc}. Related risk-control approaches also typically calibrate a risk-controlling parameter after model training is complete~\cite{Bates2021-wy,Angelopoulos2022-as}.

Our approach differs by incorporating the dismissal criterion directly into training. We introduce a closed-loop, certificate-aware training objective in which the dismissal threshold is repeatedly updated during optimization and used to guide the classifier away from unsafe dismissal of cancer-positive examinations. In contrast to conventional post-hoc thresholding, the operating criterion therefore becomes part of the learning process itself. Final safe-dismissal performance is evaluated using an independently derived statistical certificate.

This study makes three main contributions. First, we introduce a closed-loop training objective that dynamically couples dismissal-threshold estimation with model optimization. Second, we use a controlled ablation study to distinguish the effect of dynamic threshold updating from the underlying classification losses and from a fixed-threshold alternative. Third, we evaluate the method on two independent mammography datasets using internal and cross-institutional external validation, with formally certified case-level dismissal and complementary image-level analysis.

\section{Related Work}
\label{sec:related}

\subsection{Deep learning for mammography screening}

Large-scale deep learning systems have shown strong performance in breast cancer screening. Some studies have reported performance comparable to radiologists in retrospective evaluations~\cite{Wu2020-gg,Kwon2024-me}, while others have evaluated AI as an independent or integrated reader within population-based screening workflows~\cite{Van_Winkel2025-ps,Eisemann2025-ut}. Prospective studies have also demonstrated that AI can be incorporated into clinical screening workflows to support cancer detection while reducing radiologist workload~\cite{Lang2023-vd,Hernstrom2025-wf,Gommers2026-db}. These studies have primarily evaluated AI using conventional measures such as AUROC, sensitivity, and specificity, together with measures of workload reduction.

More directly related to AI-based rule-out, Bernstein et al.\ proposed a statistical framework for selecting rule-out thresholds in screening mammography by examining the trade-off between caseload reduction and the risk of missed cancers~\cite{Bernstein2026}. Their analysis provides an important framework for evaluating the consequences of different rule-out thresholds. However, the thresholds are evaluated using scores from an already trained model and do not influence the model training process itself. This distinction is important for the present study, where the dismissal criterion is incorporated directly into the model optimization stage rather than being considered only after model training.

Domain-specific pre-training has also been explored to improve mammography representation learning and downstream performance. Mammo-CLIP, for example, uses vision-language pre-training on mammography data to support downstream classification tasks~\cite{Ghosh2024-lb}. We use its pre-trained image encoder as the common backbone across all configurations in this study.

\subsection{Selective classification and conformal risk control}

Selective classification allows a model to abstain when its prediction is not sufficiently reliable. Geifman and El-Yaniv extended this idea to deep neural networks using confidence-based thresholding with finite-sample risk control~\cite{geifman2017selective}. They later introduced SelectiveNet, which jointly learns the prediction and selection functions within a single architecture~\cite{Geifman2019-yc}. Unlike these approaches, our method does not introduce a separate selection head. Instead, the dismissal boundary is used directly to guide optimization of the classifier itself.

Conformal prediction and conformal risk control provide related approaches for controlling uncertainty and predictive risk under exchangeability assumptions~\cite{Angelopoulos2021-fh,Angelopoulos2022-as,Bates2021-wy}. More recently, Yeh et al.\ introduced Conformal Risk Training, an end-to-end framework that incorporates conformal risk control into model training by differentiating through the risk-control procedure~\cite{Yeh2025CRT}. This work demonstrates the value of allowing a statistical risk criterion to influence model optimization rather than applying it only after training.

Our approach is related in motivation but focuses specifically on safe dismissal in breast cancer screening. Rather than applying risk control only after training, we incorporate a dynamically updated dismissal criterion into the training process so that the classifier is optimized for high-recall dismissal. Final case-level dismissal is then evaluated using an independent statistical certification procedure described in Section~\ref{sec:methods}.

\subsection{Calibration under class imbalance}

Strong discriminative performance does not necessarily imply well-calibrated probability estimates. Modern neural networks can be poorly calibrated and are often overconfident in their predictions~\cite{SAMBYAL2023107816}. This issue is particularly relevant in breast cancer screening, where class imbalance can affect both model optimization and the reliability of predicted scores.

Focal loss was originally introduced to reduce the influence of easily classified examples in highly imbalanced detection tasks~\cite{Lin2017-kb}. Later studies showed that, under certain conditions, it can also improve calibration compared with standard cross-entropy~\cite{Mukhoti2020-qc}. Brier-based objectives provide another approach by directly penalizing the squared difference between predicted probabilities and observed labels, and they have also been used as training objectives in medical imaging~\cite{Sander2020-gu}. These methods were therefore included in our ablation study to examine whether improvements in calibration or class-imbalance handling also translate into better safe-dismissal performance.

\subsection{Domain shift and external validation}

Medical imaging models do not always generalize well across institutions, scanner types, and acquisition protocols. In some cases, model performance can be influenced by site-specific characteristics that are unrelated to the underlying disease~\cite{zech2018variable}. Similar problems have been reported in mammography, where differences in imaging systems and acquisition conditions can reduce performance across datasets and institutions~\cite{GARRUCHO2022102386}. Recent cross-dataset work has also shown that dataset-specific patterns can remain detectable even after consistent pre-processing, and that adding external abnormal-enriched data does not necessarily improve performance on the target screening population~\cite{Hajishafiezahramini2026-gr}.

Most studies of domain shift focus on changes in conventional performance measures such as AUROC. In this study, we also examine how safe-dismissal performance changes when the model is evaluated on data from an independent institution. This allows us to assess whether the benefit of certificate-aware training is maintained under cross-institutional domain shift.

\section{Materials and Methods}
\label{sec:methods}

\subsection{Datasets}

\paragraph{NLBS.}
The Newfoundland and Labrador Breast-Screening (NLBS) dataset was obtained from the provincial breast screening program of Newfoundland and Labrador Health Services. It consists of anonymized full-field digital mammograms stored in DICOM format. The published dataset includes 26{,}988 images from 5{,}997 screening cases, of which 149 cases were cancer-positive
~\cite{Kendall2025NLBS,Kendall2025NLBSdataset}.

The 149 cancer-positive cases contain 652 images across both breasts and all available views. Since cancer was usually present in only one breast, 337 images from the positive cases were
assigned a positive image-level label. Images from the unaffected contralateral breast in the same cases were labeled negative. Mammograms were acquired using GE Senograph Essential full-field digital mammography systems manufactured between 2008 and 2010.

Cancer-positive cases were confirmed through further diagnostic procedures and laboratory analysis. Cases initially considered suspicious during screening but later shown to be nonmalignant were classified as false positive based on diagnostic imaging and/or laboratory findings. Normal cases were considered cancer-free when no interval cancer was identified during at least two years of follow-up.

\paragraph{RSNA.}
We used the publicly available training partition of the RSNA Screening Mammography Breast Cancer Detection Challenge dataset~\cite{Trivedi2026-bz}. The dataset includes screening examinations collected at Emory Healthcare in the United States and BreastScreen Victoria in Australia. The training partition contains 11{,}913 examinations, with one examination per participant, and was enriched to approximately 4\% cancer prevalence~\cite{Trivedi2026-bz}. After preprocessing, the cohort used in this study contained 54{,}706 images from 11{,}913 examinations, including 486 cancer-positive examinations, corresponding to a case-level prevalence of 4.1\%. In total, 1{,}158 images were associated with cancer-positive breasts.

Screen-detected cancer was defined as an abnormal screening examination followed by pathological confirmation of malignancy through biopsy or surgical excision. Negative examinations included those confirmed as negative or benign through follow-up, benign biopsy, or diagnostic assessment~\cite{Trivedi2026-bz}. Cancer labels were provided at the breast level rather than independently for each image.

For comparison with NLBS, the primary RSNA analysis was performed at the
examination level, with \texttt{patient\_id} defining the case. An
examination was considered cancer-positive if either breast was labeled
cancer-positive.

\subsection{Preprocessing}

Both datasets were processed with the same DICOM-to-PNG pipeline to keep preprocessing consistent across the NLBS and RSNA datasets. The modality LUT and, when available, the VOI LUT were applied to the original DICOM pixel data~\cite{dicomPS33}. MONOCHROME1 images were inverted so that breast tissue appeared brighter than the background. Each image was then min--max normalized to $[0,1]$. This normalization was used only during the initial segmentation and cropping stage. The image intensity range was recalculated later in the model data loader.

Right-breast images were horizontally mirrored so that all images had a consistent orientation. Image cropping used a low threshold based on the median intensity along the image border. Holes inside the breast mask were filled, followed by morphological closing with a $9\times9$ kernel. Each image was cropped to the breast bounding box with a 10\% margin. Pixels outside the breast mask were set to zero, and the resulting grayscale images were saved as 16-bit PNG files.

Before being passed to the network, each cropped PNG image was resized to $1520\times912$ pixels using linear interpolation, following the input size used in Mammo-CLIP~\cite{Ghosh2024-lb}. The resized image was min--max normalized to $[0,1]$. During training, data augmentation included random rotations of up to $\pm10^{\circ}$, isotropic scaling between 0.9 and 1.1, and translations of up to 5\% of the image dimensions. Brightness was varied by $\pm0.05$, and contrast was scaled between 0.90 and 1.10. Pixel values were clipped to $[0,1]$ after augmentation.

The single-channel image was then replicated across three channels and standardized using the channel-wise mean and standard deviation of the released Mammo-CLIP checkpoint~\cite{Ghosh2024-lb}. Calibration and test images were processed without random augmentation and underwent only resizing, min--max normalization, channel replication, and Mammo-CLIP standardization.

\subsection{Model architecture}
\label{sec:architecture}

We used EfficientNet-B5 as the image encoder~\cite{pmlr-v97-tan19a}, initialized with the pre-trained Mammo-CLIP weights~\cite{Ghosh2024-lb}. The encoder was kept frozen during training. Only the classification head was trained. This head consisted of LayerNorm, dropout with a probability of 0.3, and a single linear output unit. A sigmoid function was applied to the final output to obtain a score $p\in[0,1]$.

Keeping the encoder fixed allowed us to compare the different training objectives without changing the underlying image representation. Therefore, any differences between configurations were mainly due to the loss function and optimization of the classification head. The sigmoid outputs were treated as model scores rather than calibrated estimates of cancer probability, since balanced minibatch sampling changed the class distribution seen during training.

\subsection{Case-level score and label definitions}
\label{sec:aggregation}

Let $\mathcal{I}_c$ represent the set of images belonging to case $c$. The case-level prediction score was defined as the maximum score among all images in that case:

\begin{equation}
p_c = \max_{i\in\mathcal{I}_c} p_i,
\label{eq:maxaggregation}
\end{equation}

where $p_i$ is the sigmoid output for image $i$. The case-level label was defined in the same way:

\begin{equation}
y_c = \max_{i\in\mathcal{I}_c} y_i,
\end{equation}

where $y_i\in\{0,1\}$ is the image-level cancer label. A case was therefore considered cancer-positive if at least one of its images had a positive label.

Maximum aggregation was chosen as a conservative examination-level rule because a case should remain for radiologist review if any constituent image receives a high cancer score. Consequently, a case is dismissed only when all of its images fall below the operating threshold.

\subsection{Certification unit and the independence assumption}
\label{sec:independence}

The Clopper--Pearson bound in Equation~\eqref{eq:cpupper} is an exact binomial confidence bound. In our setting, this requires each observation used for certification to contribute one independent binary outcome: whether a dismissed case contains cancer or not~\cite{10.1093/biomet/26.4.404}. Multiple images from the same screening examination cannot be treated as separate independent observations because they come from the same patient and share acquisition-related characteristics. Treating these images as independent would artificially increase the sample size used for certification and could produce an overly narrow confidence bound.

For this reason, formal certification is performed only at the case level. Image scores are first combined using the maximum-aggregation rule in Section~\ref{sec:aggregation}, so that each case contributes a single score and a single binary outcome to the Clopper--Pearson calculation. The resulting confidence bound therefore applies to the cancer rate among dismissed cases, not among individual dismissed images.

Image-level dismissal results are reported only as a secondary descriptive analysis. They are used to show how the model behaves at the individual-image level, but no Clopper--Pearson confidence bound or formal certification is assigned to those results. Throughout this paper, the statistical guarantee refers only to the case-level analysis.

\subsection{Five-fold cross-validation and calibration splits}
\label{sec:splitting}

All five ablation configurations described in Section~\ref{sec:ablation} were trained using the same five-fold cross-validation scheme. Splitting was performed at the case level using \texttt{StratifiedGroupKFold}, with stratification based on case-level cancer status and grouping by case identifier. This ensured that all images belonging to the same case remained in the same fold. Fold assignments were generated using a fixed random seed and reused for every configuration so that all models were compared under identical data splits.

For each fold, one fifth of the dataset (20\%) was reserved as the outer
test partition. The remaining 80\% formed the development data and was
divided at the case level, with stratification by cancer status, into
70\% of the full dataset for model fitting and 10\% for training
calibration. The fitting subset was used for gradient updates, while the
training-calibration subset was used to recalculate the provisional
threshold $\tau_t$ during training, as described in
Section~\ref{sec:loss}. This threshold served only as a training-time
optimization signal for the closed-loop loss and was not used as the
final reported operating threshold. Thus, in each fold, 70\% of the cases
were used for fitting, 10\% for training calibration, and 20\% for outer
testing.

All configurations were trained for a fixed 20 epochs. No early stopping or validation-based checkpoint selection was used, and the final-epoch checkpoint was used for inference in every configuration. This kept the training procedure identical across the ablation study and avoided differences caused by model or epoch selection.

After training each fold, predictions were generated for its outer test partition. Repeating this process across all five folds produced one out-of-fold prediction for every case in the dataset. Each out-of-fold prediction was therefore generated by a model for which that case had not been used for model fitting or training-time threshold calculation. The five outer-test prediction sets were then pooled to form a complete cross-validated prediction set covering the full dataset. Thus, every prediction included in the subsequent threshold-search and evaluation analyses was obtained out of fold, before the final search/evaluation split was created.

The final operating-threshold search and statistical evaluation were
performed only after all out-of-fold predictions had been generated and
pooled. The pooled predictions were then divided at the case level, with
stratification by cancer status, into a 20\% threshold-search subset and a
separate 80\% evaluation subset using a fixed random seed. This downstream
split was separate from the training-calibration subset used during model
optimization and did not alter or regenerate any model predictions. The
20\% subset was used only to determine the operating thresholds for the
98\% and 95\% recall targets. The selected thresholds were then fixed and
applied once to the independent 80\% evaluation subset. No labels or
outcomes from the evaluation subset were used to select or adjust the
thresholds.
\subsection{Threshold certification and evaluation}
\label{sec:certified}

Threshold selection and final evaluation were performed after completion
of the five-fold cross-validation. The outer-test predictions from all
five folds were first pooled, giving one out-of-fold prediction for every
case in the dataset. These pooled predictions were then divided, at the
case level and stratified by cancer status, into a 20\% threshold-search
subset and a separate 80\% evaluation subset for all five ablation configurations.

For each configuration, two operating thresholds were selected on the
20\% threshold-search subset, corresponding to target cancer recalls of
98\% and 95\%. Because recall is discrete when the number of
cancer-positive cases is limited, we selected the largest threshold whose
empirical recall remained at or above the target value. The achieved
recall on the threshold-search subset could therefore exceed the nominal
target when the target value was not exactly attainable.

Once selected, each threshold was fixed and applied once to the independent
80\% evaluation subset without further adjustment. No labels or outcomes
from the evaluation subset were used to select or modify the operating
thresholds. The dismissal rate, achieved cancer recall, and corresponding
Clopper--Pearson upper confidence bound were calculated only on this
evaluation subset.

For a fixed operating threshold, let $N$ be the total number of cases in the evaluation subset, $n_d$ the number of dismissed cases, $k_d$ the number of cancer-positive cases among those dismissed, and $C$ the total number of cancer-positive cases in the evaluation subset.

The dismissal fraction was calculated as

\begin{equation}
d=\frac{n_d}{N}
\end{equation}

The achieved cancer recall was

\begin{equation}
\mathrm{Recall}=1-\frac{k_d}{C}.
\end{equation}

To account for uncertainty in the observed cancer rate among dismissed cases, we calculated a one-sided Clopper--Pearson upper confidence bound:

\begin{equation}
U_{\mathrm{dismiss}}
=
\begin{cases}
1, & n_d=0 \text{ or } k_d=n_d,\\[3pt]
\mathrm{Beta}^{-1}_{1-\delta}(k_d+1,n_d-k_d),
& 0\leq k_d<n_d,
\end{cases}
\label{eq:cpupper}
\end{equation}

where $\mathrm{Beta}^{-1}_{q}(a,b)$ is the $q$ quantile of the beta distribution~\cite{10.1093/biomet/26.4.404}. For the final evaluation, $\delta=0.01$ was used, corresponding to a one-sided 99\% confidence level. The resulting value gives an upper confidence limit on the cancer prevalence within the dismissed group. Formal confidence bounds were calculated only at the case level, as discussed in Section~\ref{sec:independence}.

\subsection{Proposed closed-loop dismissal loss}
\label{sec:loss}

The proposed training objective combines binary cross-entropy, focal loss~\cite{Lin2017-kb}, and a one-sided dismissal loss:

\begin{align}
\mathcal{L}
&=
\mathcal{L}_{\mathrm{CE}}
+
\lambda_{\mathrm{focal}}\mathcal{L}_{\mathrm{focal}}
+
\lambda_{\mathrm{dismiss}}\mathcal{L}_{\mathrm{dismiss}},
\\
\mathcal{L}_{\mathrm{dismiss}}
&=
\frac{1}{N_{C+}}
\sum_{i:y_i=1}
\max\left(0,\tau_t+m-p_i\right),
\end{align}

where $N_{C+}$ is the number of cancer-positive images in the minibatch, $p_i$ is the predicted score for image $i$, $m$ is a fixed margin, and $\tau_t$ is the provisional dismissal threshold at epoch $t$.

The dismissal term is applied only to cancer-positive examples. A positive example receives an additional penalty when its score falls below $\tau_t+m$, and the penalty increases as the score moves farther below this boundary. Positive examples with scores above the boundary receive no dismissal penalty. The cross-entropy and focal-loss terms continue to provide the classification signal for both positive and negative examples.

For the proposed closed-loop model, $\tau_t$ was recalculated at the
beginning of each epoch using image-level predictions from that fold's
training-calibration subset described in Section~\ref{sec:splitting}.
We selected the largest threshold for which the one-sided 95\%
Clopper--Pearson upper bound on the positive-image rate among dismissed
calibration images did not exceed 1\%. If no threshold satisfied this
criterion, $\tau_t$ was set to zero. This calculation was used only as a
training-time signal and was separate from the final 99\% case-level
evaluation described in Section~\ref{sec:certified}. No gradient was
propagated through the threshold calculation.

\begin{figure}[htbp]
    \centering
    \includegraphics[width=\textwidth] {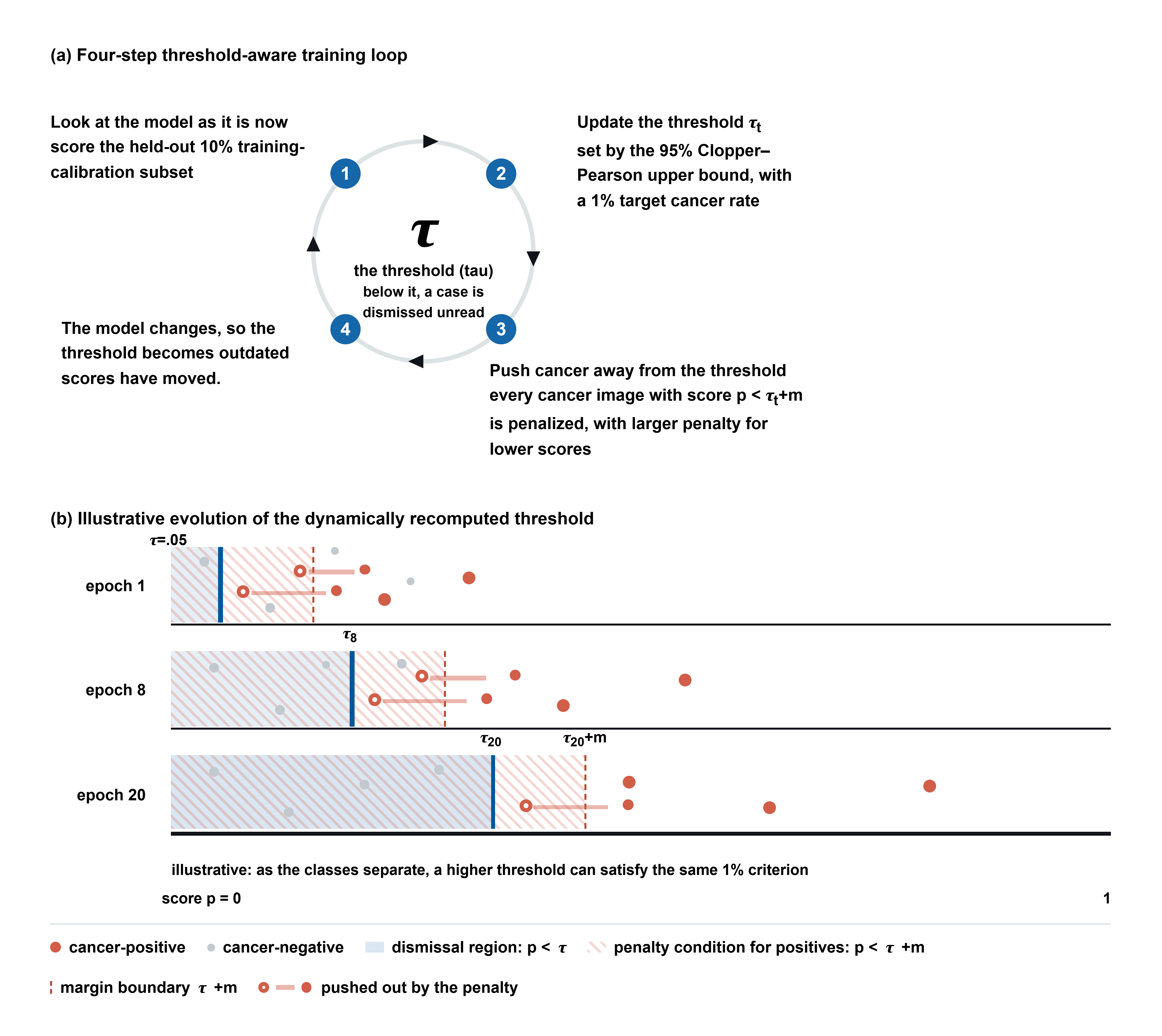}
    \caption{
    Closed-loop threshold-aware training.
    (a) Four-step training cycle in which the provisional threshold $\tau_t$
    is recomputed from the training-calibration subset and used to penalize
    cancer-positive images with scores below $\tau_t+m$.
    (b) Illustrative evolution of $\tau_t$ as the model score distribution
    changes during training. The blue region denotes the dismissal region
    ($p<\tau_t$), and the hatched region denotes the positive-sample penalty
    condition ($p<\tau_t+m$). Hollow and solid cancer-positive markers
    illustrate score movement under the dismissal penalty. Threshold evolution
    is illustrative and need not be monotonic.
    }
    \label{fig:closed_loop}
\end{figure}

Figure~\ref{fig:closed_loop} illustrates the proposed closed-loop training mechanism.

The fixed hyperparameters were $\gamma=2.0$, $\lambda_{\mathrm{focal}}=1.0$, 
$\lambda_{\mathrm{dismiss}}=0.05$, and $m=0.10$. The margin was used to 
maintain a buffer between cancer-positive scores and the current dismissal 
boundary, while the relatively small dismissal weight allowed this term to 
act as an auxiliary training signal alongside the cross-entropy and focal-loss 
objectives.

The threshold variable was initialized at 0.05. In the closed-loop 
configuration, it was replaced by the value recalculated from the 
training-calibration subset at the beginning of each epoch. In the 
fixed-$\tau$ ablation, $\tau_t$ remained at 0.05 throughout training. This 
comparison was used to separate the effect of dynamically updating the 
threshold from the effect of adding the dismissal loss itself.

\subsection{Training protocol}
\label{sec:training}

All five configurations used the same data splits and training settings to ensure a fair comparison. The linear classification head was optimized using AdamW with a learning rate of $3\times10^{-5}$ and a weight decay of $10^{-4}$. Models were trained for 20 epochs with a minibatch size of 80, automatic mixed precision, and gradient clipping at a global norm of 5.0. No early stopping or validation-based checkpoint selection was used.

Because cancer-positive examples were much less common than negative examples, each training minibatch was constructed with 20 cancer-positive and 60 cancer-negative samples. Samples were drawn independently for each minibatch, allowing minority-class examples to be reused across minibatches as needed.
Since balanced sampling changes the class distribution seen during training, the sigmoid outputs were treated as model scores rather than as population-calibrated probabilities of cancer.
\subsection{Ablation configurations}
\label{sec:ablation}

Five training configurations were compared using the same data splits, training schedule, and evaluation procedure:

\begin{enumerate}

\item \textbf{CE-only baseline}: binary cross-entropy was used as the sole
training objective,
\begin{equation}
\mathcal{L}_{\mathrm{CE}}
=
-\frac{1}{N}
\sum_{i=1}^{N}
\left[
y_i\log(p_i)
+
(1-y_i)\log(1-p_i)
\right].
\end{equation}
This configuration served as the primary baseline.

\item \textbf{CE + Brier}: binary cross-entropy was combined with
Brier-score regularization,
\begin{equation}
\mathcal{L}
=
\mathcal{L}_{\mathrm{CE}}
+
\lambda_{\mathrm{Brier}}
\mathcal{L}_{\mathrm{Brier}},
\end{equation}
where
\begin{equation}
\mathcal{L}_{\mathrm{Brier}}
=
\frac{1}{N}
\sum_{i=1}^{N}
(p_i-y_i)^2,
\end{equation}
and $\lambda_{\mathrm{Brier}}=0.1$. This configuration was included to
test whether a calibration-oriented training objective also improves
safe-dismissal performance.

\item \textbf{CE + Focal}: binary cross-entropy was combined with focal
loss,
\begin{equation}
\mathcal{L}
=
\mathcal{L}_{\mathrm{CE}}
+
\lambda_{\mathrm{focal}}
\mathcal{L}_{\mathrm{focal}},
\end{equation}
where
\begin{equation}
\mathcal{L}_{\mathrm{focal}}
=
-\frac{1}{N}
\sum_{i=1}^{N}
\left[
y_i(1-p_i)^{\gamma}\log(p_i)
+
(1-y_i)p_i^{\gamma}\log(1-p_i)
\right].
\end{equation}
We used $\gamma=2.0$ and $\lambda_{\mathrm{focal}}=1.0$.

\item \textbf{Fixed-$\tau$ dismissal}: the full classification and
dismissal objective was used,
\begin{equation}
\mathcal{L}
=
\mathcal{L}_{\mathrm{CE}}
+
\lambda_{\mathrm{focal}}\mathcal{L}_{\mathrm{focal}}
+
\lambda_{\mathrm{dismiss}}\mathcal{L}_{\mathrm{dismiss}},
\end{equation}
with
\begin{equation}
\mathcal{L}_{\mathrm{dismiss}}
=
\frac{1}{N_{C+}}
\sum_{i:y_i=1}
\max(0,\tau_t+m-p_i).
\end{equation}
For this configuration, the dismissal threshold was fixed at
$\tau_t=0.05$ throughout training.

\item \textbf{Proposed closed-loop}: the same full objective was used,
but $\tau_t$ was recomputed from the training-calibration subset at the
beginning of every epoch. The fixed hyperparameters were
$\gamma=2.0$, $\lambda_{\mathrm{focal}}=1.0$,
$\lambda_{\mathrm{dismiss}}=0.05$, and $m=0.10$.

\end{enumerate}

All five configurations were retained in the final comparison. None was selected or excluded based on test-set performance, and all were evaluated using the same threshold-certification and outer-test procedure described in Section~\ref{sec:certified}.

\subsection{External evaluation}
\label{sec:externalvalidation}

For the external evaluation, one final CE-only model and one final
closed-loop model were trained on RSNA and then applied to the external
NLBS dataset without retraining the model weights. The external
analysis focused on the primary CE-only baseline and the proposed
closed-loop model, while the remaining configurations were evaluated in
the controlled in-domain ablation study. For these final models, 90\% of
the RSNA cases were used for model fitting and the remaining 10\% were used
as the training-calibration subset, following the same training procedure
and hyperparameters used in the in-domain experiments.

To keep the operating-point evaluation consistent across datasets, the same
threshold-selection procedure used for the in-domain analyses was applied
to the external evaluation. Cases were divided at the case level, with
stratification by cancer status, into a 20\% threshold-search subset and an
independent 80\% evaluation subset.

For each recall target, the operating threshold was selected using only the
20\% threshold-search subset. Because recall is discrete when the number of
cancer-positive cases is limited, the largest threshold whose empirical
recall remained at or above the target value was selected. The threshold
was then fixed and applied once to the independent 80\% evaluation subset.
No labels or outcomes from the evaluation subset were used to select or
modify the operating threshold.

The model weights remained entirely RSNA-trained. Because prediction-score
distributions may differ across datasets, the operating threshold for the
external NLBS analysis was determined from the NLBS threshold-search subset
rather than transferring an RSNA-derived threshold unchanged. Dismissal
rate, achieved cancer recall, case-level AUROC, and the one-sided 99\%
Clopper--Pearson upper confidence bound were calculated only on the
independent NLBS evaluation subset.

This experiment therefore evaluates cross-dataset performance with
target-domain threshold calibration. It tests whether the behavior learned
from RSNA is retained on an independent screening population when the model
weights remain fixed and only the operating threshold is determined from a
separate subset of the target dataset.

\subsection{Evaluation metrics}
\label{sec:metrics}

Model discrimination was evaluated using the area under the
receiver-operating characteristic curve (AUROC) at both the image and case
levels. Case-level scores were obtained using the maximum-aggregation rule
described in Section~\ref{sec:aggregation}. AUROC was calculated on the
independent 80\% evaluation subset.

Safe-dismissal performance was evaluated at cancer-recall targets of 98\%
and 95\%. For each target, the operating threshold was selected using only
the 20\% threshold-search subset, fixed, and then applied once to the
independent 80\% evaluation subset. We calculated the dismissal rate, achieved cancer recall, and the corresponding
one-sided 99\% Clopper--Pearson upper confidence bound on the independent
evaluation subset. A dismissal result is reported at a given recall target only
when the achieved recall on the evaluation subset met or exceeded that target;
otherwise, the result is reported as N/A.

Formal certification was performed only at the case level. Image-level
dismissal rates were reported as a secondary descriptive analysis and were
not assigned a Clopper--Pearson confidence bound, as discussed in
Section~\ref{sec:independence}.

All reported case-level evaluation metrics were calculated only from the
80\% evaluation subset. The 20\% threshold-search subset was used only to
select the operating thresholds and was not included in the reported
performance estimates. For the descriptive image-level analysis, the same
operating thresholds selected for the corresponding case-level recall
targets were applied directly to the individual image scores; no separate
image-level threshold selection was performed.

\section{Results}
\label{sec:results}

Before presenting the results, we briefly clarify the interpretation of the
Clopper--Pearson bound used throughout this section. For a given operating
threshold, the dismissed group contains the cases that would not be sent for
radiologist review. The one-sided 99\% Clopper--Pearson upper bound (CPU99)
represents an upper confidence bound on the cancer prevalence within this
dismissed group. For example, a CPU99 value of 1.00\% indicates that the
one-sided 99\% upper confidence limit on the cancer prevalence among dismissed
cases is 1.00\%. This quantity refers only to the dismissed group, not to the
full cohort or to the proportion of all cancer cases that were missed. A lower
CPU99 therefore represents a tighter upper confidence bound on cancer
prevalence among dismissed cases. Because separate operating thresholds were
selected for the 98\% and 95\% recall targets, the dismissed groups and their
corresponding CPU99 bounds can differ between the two operating points.

\subsection{In-domain ablation results: NLBS}
\label{sec:nlbs_results}

Table~\ref{tab:nlbs} summarizes the NLBS results across the five training
configurations. At the operating threshold selected for the 98\% recall
target, the proposed closed-loop model achieved the highest case-level
dismissal rate, at 19.74\%, together with the lowest CPU99 bound, at 0.70\%.
At the 95\% recall-target operating point, it again achieved the highest
case-level dismissal rate, at 21.70\%, with a CPU99 bound of 0.81\%.
The same overall pattern was observed at the image level, where the proposed
model achieved the highest dismissal rates at both operating thresholds,
14.80\% at $\tau_{98}$ and 30.88\% at $\tau_{95}$. The proposed model also
had the highest case-level AUROC on NLBS, at 0.7216.

\begin{table*}[htbp]
\centering

\caption{NLBS in-domain ablation results at the case and image levels.
The independent 80\% evaluation subset contained 4{,}797 cases, including
119 cancer-positive cases. The one-sided 99\%
Clopper--Pearson upper bound (CPU99) is reported only for case-level
results. Image-level dismissal rates were calculated using images from the
same 80\% evaluation subset and the same thresholds selected on the
corresponding 20\% threshold-search subset; no separate image-level
threshold search was performed.}

\label{tab:nlbs}
\small
\resizebox{\textwidth}{!}{
\begin{tabular}{lccccccc}
\toprule
Configuration &
Case @98\% target &
CPU99 &
Case @95\% target &
CPU99 &
Case AUROC &
Image dismissal at $\tau_{98}$ &
Image dismissal at $\tau_{95}$ \\
\midrule
CE-only (baseline) &
N/A  & --- &
N/A  & --- &
0.7157 &
12.08\% &
12.21\% \\
CE + Brier &
12.07\% & 1.14\% &
12.53\% & 1.39\% &
0.7148 &
12.16\% &
12.26\% \\
CE + Focal &
10.92\% & 1.26\% &
16.36\% & 1.47\% &
0.7146 &
7.70\% &
15.45\% \\
Fixed-$\tau$ dismissal &
15.45\% & 0.89\% &
18.55\% & 1.12\% &
0.7160 &
12.29\% &
17.59\% \\
\textbf{Proposed (closed-loop)} &
\textbf{19.74\%} & \textbf{0.70\%} &
\textbf{21.70\%} & \textbf{0.81\%} &
\textbf{0.7216} &
\textbf{14.80\%} &
\textbf{30.88\%} \\
\bottomrule
\end{tabular}
}
\end{table*}

\subsection{In-domain ablation results: RSNA}
\label{sec:rsna_results}

The RSNA results are shown in Table~\ref{tab:rsna}. As on NLBS, the
proposed closed-loop model achieved the highest case-level dismissal rate
at both recall-target operating points. At the threshold selected for the
98\% recall target, the proposed model dismissed 14.31\% of cases and
produced the lowest CPU99 bound, at 1.06\%. At the threshold selected for
the 95\% recall target, it dismissed 19.69\% of cases with a CPU99 bound of
1.28\%. The proposed model also achieved the highest image-level dismissal
rates at both corresponding operating thresholds.

\begin{table*}[htbp]
\centering
\caption{RSNA in-domain ablation results at the case and image levels.
The independent 80\% evaluation subset contained 9{,}530 cases, including
389 cancer-positive cases. The one-sided 99\%
Clopper--Pearson upper bound (CPU99) is reported only for case-level
results. Image-level dismissal rates were calculated using images from the
same 80\% evaluation subset and the same thresholds selected on the
corresponding 20\% threshold-search subset; no separate image-level
threshold search was performed.}

\label{tab:rsna}
\small
\resizebox{\textwidth}{!}{
\begin{tabular}{lccccccc}
\toprule
Configuration &
Case @98\% target &
CPU99 &
Case @95\% target &
CPU99 &
Case AUROC &
Image dismissal at $\tau_{98}$ &
Image dismissal at $\tau_{95}$ \\
\midrule
CE-only (baseline) &
7.04\% & 2.37\% &
13.49\% & 1.77\% &
0.7219 &
10.51\% &
16.57\% \\
CE + Brier &
8.96\% & 1.86\% &
13.42\% & 1.98\% &
0.7215 &
10.35\% &
16.49\% \\
CE + Focal &
7.48\% & 2.23\% &
17.82\% & 1.49\% &
0.7267 &
7.10\% &
21.08\% \\
Fixed-$\tau$ dismissal &
9.91\% & 1.69\% &
17.10\% & 1.56\% &
0.7280 &
8.81\% &
17.12\% \\
\textbf{Proposed (closed-loop)} &
\textbf{14.31\%} & \textbf{1.06\%} &
\textbf{19.69\%} & \textbf{1.28\%} &
0.7136 &
\textbf{11.03\%} &
\textbf{23.24\%} \\
\bottomrule
\end{tabular}
}
\end{table*}

One difference from NLBS was the AUROC ranking. On RSNA, the proposed
model had a case-level AUROC of 0.7136, while the other configurations
ranged from 0.7215 to 0.7280. Despite these relatively similar AUROC
values, the dismissal results differed more clearly across configurations,
with the proposed model achieving the highest dismissal rate and the lowest
CPU99 bound at both recall targets. This shows that AUROC and dismissal
performance are not necessarily aligned, particularly when performance is
evaluated at specific high-recall operating points.

\subsection{External validation: RSNA $\to$ NLBS}
\label{sec:external_results}

To evaluate performance across datasets, models trained on RSNA were applied
to NLBS without retraining the model weights. The operating thresholds were
selected on the NLBS search subset and then applied to the separate NLBS
evaluation subset, following the procedure described in
Section~\ref{sec:externalvalidation}. Table~\ref{tab:external} summarizes
the case-level results.

\begin{table}[htbp]
\centering
\caption{RSNA$\to$NLBS external evaluation at the case level.
Both models were trained exclusively on RSNA. Operating thresholds were
selected using the 20\% NLBS threshold-search subset and then fixed and
applied once to the independent 80\% NLBS evaluation subset. CPU99 denotes
the one-sided 99\% Clopper--Pearson upper confidence bound on cancer
prevalence within the dismissed group.}
\label{tab:external}
\small

\resizebox{\textwidth}{!}{
\begin{tabular}{lccccc}
\toprule
Configuration &
Dismissal @98\% target &
CPU99 (@98\%) &
Dismissal @95\% target &
CPU99 (@95\%) &
AUROC \\
\midrule

Baseline (CE-only) &
N/A &
N/A &
13.55\% &
1.77\% &
0.7184 \\

\textbf{Proposed (closed-loop)} &
\textbf{12.95\%} &
\textbf{1.06\%} &
\textbf{19.87\%} &
\textbf{1.37\%} &
0.7210 \\

\bottomrule
\end{tabular}
}

\end{table}

At the 98\% recall target, the CE-only baseline did not reach the required
recall level. The proposed closed-loop model reached the target while
dismissing 12.95\% of cases, with a CPU99 bound of 1.06\%.

At the 95\% recall target, both models reached the required recall level.
The proposed model dismissed 19.87\% of cases, compared with 13.55\% for the
baseline, an increase of 6.32 percentage points. Its CPU99 bound was also
lower, at 1.37\% compared with 1.77\% for the baseline, indicating a tighter
upper confidence bound on the cancer prevalence within the dismissed group.

Case-level AUROC was similar for the two models, with 0.7210 for the proposed
model and 0.7184 for the baseline. Overall, the proposed model maintained its
dismissal advantage when transferred from RSNA to NLBS with target-domain
threshold calibration. It also reached the stricter 98\% recall target,
which the baseline did not reach under the same evaluation procedure.

\subsection{Summary across settings}
\label{sec:summary_results}

Table~\ref{tab:master} summarizes the case-level dismissal results across
the two in-domain evaluations and the external RSNA$\to$NLBS evaluation.
For the in-domain datasets, the proposed model is compared with the
non-proposed configuration that achieved the highest dismissal rate at
each recall-target operating point. For the external evaluation, the
comparison is made against the RSNA-trained CE-only baseline.

\begin{table*}[htbp]
\centering
\caption{Summary of case-level dismissal rates and one-sided 99\%
Clopper--Pearson upper confidence bounds (CPU99) across the three
evaluation settings. For the in-domain comparisons, the best alternative
is the non-proposed configuration with the highest dismissal rate at the
corresponding recall-target operating point. For RSNA, fixed-$\tau$ is
the best alternative at the 98\% recall target and CE~+~Focal at the
95\% recall target.}
\label{tab:master}
\small
\begin{tabular}{llcccc}
\toprule
Setting & Configuration & @98\% target & CPU99 & @95\% target & CPU99 \\
\midrule

\multirow{2}{*}{NLBS in-domain}
& Best alternative (Fixed-$\tau$)
& 15.45\% & 0.89\%
& 18.55\% & 1.12\% \\

& \textbf{Proposed}
& \textbf{19.74\%} & \textbf{0.70\%}
& \textbf{21.70\%} & \textbf{0.81\%} \\

\midrule

\multirow{2}{*}{RSNA in-domain}
& Best alternative
& 9.91\% & 1.69\%
& 17.82\% & 1.49\% \\

& \textbf{Proposed}
& \textbf{14.31\%} & \textbf{1.06\%}
& \textbf{19.69\%} & \textbf{1.28\%} \\

\midrule

\multirow{2}{*}{RSNA$\to$NLBS external}
& Baseline (CE-only)
& N/A & ---
& 13.55\% & 1.77\% \\

& \textbf{Proposed}
& \textbf{12.95\%} & \textbf{1.06\%}
& \textbf{19.87\%} & \textbf{1.37\%} \\

\bottomrule
\end{tabular}
\end{table*}

Across both in-domain datasets, the proposed closed-loop model achieved
the highest case-level dismissal rate at the operating thresholds selected
for the 98\% and 95\% recall targets. On NLBS, dismissal reached 19.74\%
and 21.70\%, compared with 15.45\% and 18.55\% for the strongest
alternatives at the corresponding operating points. The associated CPU99
bounds were also lower for the proposed model, decreasing from 0.89\% to
0.70\% at the 98\% recall-target operating point and from 1.12\% to
0.81\% at the 95\% recall-target operating point.

On RSNA, the proposed model dismissed 14.31\% of cases at the 98\%
recall-target operating point, compared with 9.91\% for the strongest
alternative, while CPU99 decreased from 1.69\% to 1.06\%. At the 95\%
recall-target operating point, dismissal increased from 17.82\% to
19.69\%, while CPU99 decreased from 1.49\% to 1.28\%.

The external RSNA$\to$NLBS evaluation showed the same overall pattern.
At the 95\% recall target, the proposed model dismissed 19.87\% of cases
compared with 13.55\% for the CE-only baseline, while CPU99 decreased
from 1.77\% to 1.37\%. At the stricter 98\% recall target, the proposed
model reached the required operating point with 12.95\% dismissal and a
CPU99 of 1.06\%, whereas the baseline did not reach the target.

AUROC did not follow the same ordering across datasets. The proposed
model had the highest AUROC on NLBS, the lowest among the five
configurations on RSNA, and a slightly higher AUROC than the baseline in
the external evaluation. This indicates that AUROC and dismissal
performance are not necessarily aligned. Despite these differences in
AUROC, the proposed model achieved the highest case-level dismissal rate
at both recall-target operating points in the two in-domain evaluations
and also achieved higher dismissal than the CE-only baseline in the
external evaluation. It also achieved the highest reported image-level
dismissal rate at both corresponding operating thresholds in the two
in-domain analyses.

\section{Discussion} 
\label{sec:discussion}

\subsection{Evidence in support of the main claim}

The main finding of this study is the consistency of the dismissal
improvement across datasets and operating points. On both NLBS and RSNA,
the proposed closed-loop model achieved the highest case-level dismissal
rate at the 98\% and 95\% recall targets, while also producing the lowest
CPU99 bound at each operating point. This pattern was observed despite
differences between the two datasets in prevalence, population, and image
acquisition.

The comparison with the fixed-$\tau$ ablation is particularly informative
because the two configurations use the same dismissal loss and differ
primarily in whether the threshold is updated during training. At the
98\% recall target, closed-loop recomputation increased dismissal from
15.45\% to 19.74\% on NLBS, a gain of 4.29 percentage points, and from
9.91\% to 14.31\% on RSNA, a gain of 4.40 percentage points. The same
direction was observed at the 95\% recall target, where dismissal increased
from 18.55\% to 21.70\% on NLBS and from 17.10\% to 19.69\% on RSNA.
Importantly, these increases were not obtained at the expense of a weaker
statistical bound. CPU99 also decreased at each of these comparisons.

Taken together, these results suggest that the benefit does not come only
from adding a dismissal-oriented penalty to the loss. Updating the
dismissal boundary during training appears to provide an additional
advantage over keeping that boundary fixed. The consistency of this
difference across two datasets and two recall targets is more informative
than any single percentage-point gain.

The AUROC results provide an additional perspective. On NLBS, the proposed
model had the highest AUROC, whereas on RSNA it had the lowest AUROC among
the five configurations, with all models remaining within a relatively
narrow range. Yet the proposed model achieved the strongest dismissal
performance on both datasets. This suggests that overall ranking
performance, as summarized by AUROC, and performance at a specific
high-recall dismissal operating point are related but not interchangeable.
For the intended use considered here, the latter is the quantity that
directly determines how many examinations can be dismissed while
maintaining the required cancer recall.

\subsection{Consistency between case- and image-level results}

The case- and image-level results showed a consistent pattern across both
datasets. The proposed model achieved the highest dismissal rate at both
recall targets not only after case-level aggregation, but also when the
images were evaluated individually. On NLBS, image-level dismissal reached
14.80\% at $\tau_{98}$ and 30.88\% at $\tau_{95}$. On RSNA, the
corresponding rates were 11.03\% and 23.24\%. These were the highest
image-level dismissal rates among the five configurations at each operating
point.

This consistency is important because the primary analysis uses
maximum aggregation across the images belonging to a case. An improvement
seen only after aggregation could raise the possibility that the result
depends strongly on the aggregation rule. Instead, the same overall
advantage was observed before aggregation at the image level. The findings
therefore suggest that the improvement is not limited to the case-level
decision rule, but is also reflected in the underlying image-level score
behavior.

The image-level analysis remains descriptive, since the formal
Clopper--Pearson confidence bound is applied only at the case level.
Nevertheless, the agreement between the two levels provides additional
support that the observed dismissal advantage is not an artifact of
case-level aggregation alone.

\subsection{Discrimination and dismissal performance capture different
aspects of model behavior}

The relationship between AUROC and dismissal performance was not consistent
across the three evaluation settings. On NLBS, the proposed model achieved
the highest case-level AUROC as well as the highest dismissal rates. On
RSNA, however, it had the lowest AUROC among the five configurations while
still achieving the highest dismissal rate and the lowest CPU99 bound at
both recall targets. In the external RSNA$\to$NLBS evaluation, AUROC was
similar for the proposed model and the CE-only baseline, while the proposed
model showed a clear advantage in dismissal performance.

These results suggest that AUROC alone does not fully describe performance
for the dismissal task considered here. AUROC summarizes how well a model
ranks positive and negative cases across the full range of possible
thresholds. In contrast, the dismissal analysis evaluates the model at
specific high-recall operating points, where the practical question is how
many cases can be dismissed while maintaining the required cancer recall
and controlling the cancer rate within the dismissed group. Models with
similar AUROC values can therefore behave differently at these particular
operating points.

This distinction is especially relevant to the objective used in the
proposed model. The dismissal term does not directly optimize global
ranking performance. Instead, it penalizes cancer-positive examples that
fall too close to the current dismissal boundary. Its effect may therefore
be reflected more clearly in dismissal performance than in a threshold-free
summary measure such as AUROC. The RSNA results illustrate this point most
clearly: the proposed model did not improve AUROC, yet it achieved both a
higher dismissal rate and a lower CPU99 bound.

More broadly, clinical prediction studies have emphasized that different
performance measures describe different properties of a model and should
not be treated as interchangeable
~\cite{VanCalster2019calibration,Huang2020calibration}. Our results show a
similar distinction between overall discrimination and performance at a
specific safety-sensitive operating point. For a system intended to support
safe dismissal, both remain important, but AUROC by itself is not sufficient
to characterize how the model behaves under the intended decision rule.

\subsection{External validation under domain shift}

The external RSNA$\to$NLBS evaluation provides encouraging evidence that
the dismissal advantage is not limited to the dataset on which the model
was trained. The model weights were learned entirely from RSNA and were
kept fixed when the models were evaluated on NLBS. To account for the
change in data distribution, the operating thresholds for both models
were selected using the NLBS threshold-search subset and then applied
without further adjustment to the separate NLBS evaluation subset.

The difference between the two models was most apparent at the stricter
98\% recall target. The proposed model reached this operating point while
dismissing 12.95\% of cases, with a CPU99 bound of 1.06\%. The CE-only
baseline did not reach the required recall target under the same
evaluation procedure. At the 95\% recall target, both models reached the
target, but the proposed model dismissed 19.87\% of cases compared with
13.55\% for the baseline. The corresponding CPU99 bound was also lower for
the proposed model, at 1.37\% compared with 1.77\%.

These results are notable because the advantage was retained even though
the model parameters had not been fitted to the NLBS population. Both
models were given the same opportunity to calibrate their operating
threshold on target-domain data, so the comparison does not depend on
applying an RSNA-derived threshold unchanged to a different population.
Instead, it asks whether the advantage learned during training remains
after both models are adapted to the new dataset only through threshold
selection. Under this setting, the closed-loop model continued to dismiss
more cases while maintaining the required recall level.

The external AUROC values were similar, at 0.7210 for the proposed model
and 0.7184 for the baseline, while their dismissal performance differed
more clearly. This is consistent with the in-domain findings and further
shows that similar overall discrimination does not necessarily lead to
similar performance at a specific high-recall operating point.

The result should nevertheless be interpreted as evidence of
cross-dataset robustness rather than proof of universal generalizability.
Only one external target dataset was evaluated, and the operating
threshold was recalibrated using target-domain data before final
evaluation. Further validation across institutions, acquisition systems,
and screening populations would be needed to determine how consistently
the observed advantage transfers to other clinical settings.

\subsection{Clinical relevance and intended role}

The clinical role considered in this study is selective workload reduction,
rather than replacement of radiologist interpretation. The model is used to
identify a low-score subset of screening examinations that may be eligible
for dismissal while maintaining a predefined cancer-recall requirement.
The accompanying CPU99 bound provides an additional measure of uncertainty
by placing a one-sided upper confidence bound on the cancer prevalence
within that dismissed group.

This distinction is important when interpreting the results. The dismissal
rates reported here should not be compared directly with radiologist
sensitivity, false-negative rates, or other measures calculated over the
entire screening population. These quantities answer different questions.
Our analysis focuses specifically on the composition of the subset selected
for dismissal and on how large that subset can become while preserving the
required recall level.

Previous studies, including the MASAI trial
~\cite{Lang2023-vd,Hernstrom2025-wf,Gommers2026-db} and the international
evaluation by McKinney et al.~\cite{McKinney2020-zk}, have shown that AI can
contribute to screening workflows in ways that reduce radiologist workload.
We view the present method as complementary to this broader direction.
Rather than proposing a complete clinical reading system, it provides a
model-level framework for defining and auditing a dismissal operating point
under an explicit statistical criterion.

The magnitude of the observed dismissal rates suggests that this approach
could translate into a meaningful reduction in the number of examinations
requiring review. At the stricter 98\% recall target, the proposed model
dismissed 19.74\% of cases in the NLBS in-domain evaluation and 14.31\% on
RSNA. In the external RSNA$\to$NLBS evaluation, the RSNA-trained model
continued to dismiss 12.95\% of cases after target-domain threshold
calibration while reaching the same recall target. At the 95\% recall
target, the corresponding dismissal rates increased further.

These results establish the potential of the method as a selective
screening component rather than as a stand-alone diagnostic system. Its
practical value lies in identifying a substantial subset of examinations
for which review may potentially be avoided under a predefined operating
criterion, while keeping the decision rule and its statistical uncertainty
explicit and auditable. Prospective workflow studies would be the next step
to determine how this reduction translates into radiologist workload,
screening efficiency, and clinical outcomes in practice.

\subsection{Potential economic implications of selective dismissal}

The NLBS results also provide an indication of the potential economic scale
of selective dismissal in population-based screening. At the stricter
operating point selected for a cancer-recall target of at least 98\%, TRUST
dismissed 19.74\% of screening examinations from manual review. This
corresponds to approximately 197 fewer radiologist interpretations for every
1,000 screening examinations. Importantly, this represents a reduction in
manual interpretation rather than avoidance of the mammographic examination
itself, since image acquisition and the associated technical and
administrative costs would still occur.

As an illustrative estimate, the Ontario physician fee schedule assigns a
professional interpretation component of approximately CAD\$31 to a bilateral
screening mammogram~\cite{OntarioPhysicianSchedule2023}. Applying this unit
cost to the NLBS dismissal rate gives a gross professional interpretation-cost
equivalent of approximately CAD\$6,100 per 1,000 screening examinations.
The population-level scale could be considerably larger. For example, in a
hypothetical screening program serving 10 million women aged 40 years or
older, if each woman underwent one screening examination during a screening
round, a 19.74\% dismissal rate would correspond to approximately 1.97
million examinations not requiring manual radiologist interpretation. Using
the same illustrative professional fee, this represents approximately
CAD\$61.2 million in gross interpretation costs per screening round. These
figures are intended to illustrate the potential economic scale of workload
reduction rather than realized health-system savings. Actual savings would
depend on screening participation and frequency, local reimbursement models,
workflow implementation, and the costs of deploying and maintaining the AI
system. Importantly, this estimate considers only the professional
interpretation component and does not include any potential additional savings
from fewer recalls, supplemental imaging examinations, benign biopsies, or
other downstream diagnostic procedures. Such downstream effects could further
increase the economic benefit of selective dismissal, but they were not
quantified in the present study.

\subsection{Why dynamic threshold recomputation may matter}

The comparison between the fixed-$\tau$ and closed-loop configurations
provides some insight into why updating the dismissal threshold during
training may be useful. The score distribution of a model is not fixed
while the classifier is being optimized. As training progresses, the
relative positions of positive and negative examples can shift, meaning
that a dismissal boundary chosen at the beginning of training may no
longer represent the operating region that the model currently produces.

With a fixed threshold, the dismissal loss therefore continues to act
relative to the same numerical boundary throughout training. In the
closed-loop model, the provisional threshold is recalculated from the
current model outputs and is then used to define the positive-side
dismissal penalty for the next stage of optimization. This creates a
simple feedback mechanism in which the training signal remains tied to
the model's evolving score distribution rather than to a threshold that
may become progressively less representative.

The ablation results are consistent with this interpretation. Replacing
the fixed threshold with the dynamically recomputed threshold improved
case-level dismissal on both NLBS and RSNA and at both recall targets,
while also producing lower CPU99 bounds. Because the two configurations
otherwise share the same dismissal-loss structure, this comparison
suggests that adapting the boundary during training contributes beyond
the effect of introducing the dismissal penalty alone.

These results do not establish the exact mechanism by which the score
distribution changes during training. A more detailed analysis of score
trajectories, threshold evolution, and class-conditional score
distributions would be needed to demonstrate that mechanism directly.
Nevertheless, the consistency of the fixed-$\tau$ comparison across both
datasets supports the practical value of keeping the dismissal objective
connected to the model's current operating behavior.

\subsection{Limitations}
\label{sec:limitations}

Several limitations should be considered when interpreting these results.

\begin{itemize}

\item \textbf{Limited number of cancer-positive cases in NLBS.}
Although NLBS contains 5{,}997 screening examinations, only 149 are
cancer-positive. This limits the resolution of recall-based threshold
selection, particularly within the 20\% threshold-search subset, and makes
the selected operating threshold more sensitive to individual positive
cases. The agreement between NLBS and the larger RSNA dataset is therefore
important, but additional evaluation in larger screening cohorts would
provide more stable threshold estimates.

\item \textbf{Sensitivity to loss hyperparameters was not systematically
evaluated.}
The margin $m=0.10$ was introduced to maintain a buffer between
cancer-positive scores and the current dismissal boundary, while
$\lambda_{\mathrm{dismiss}}=0.05$ was chosen to keep the dismissal term
as an auxiliary component of the overall training objective. These values
were fixed across both datasets and were not optimized separately for
individual experiments. Although the same setting produced consistent
results on NLBS and RSNA, a broader sensitivity analysis would be useful
to determine how strongly dismissal performance depends on these
hyperparameters.

\item \textbf{Comparison with established selective-classification
methods is limited.}
The current experiments compare the proposed method with controlled
ablations of the same model and training pipeline. We did not directly
compare it with dedicated selective-classification approaches such as
SelectiveNet~\cite{Geifman2019-yc} or with conformal risk-control methods
\cite{Angelopoulos2022-as} under matched operating conditions. Such
comparisons would help clarify how much benefit is provided by the proposed
closed-loop training strategy relative to alternative approaches to risk
control and selective prediction.

\item \textbf{A single backbone and pre-training source were used.}
All experiments used a frozen EfficientNet-B5 encoder initialized from the
same Mammo-CLIP checkpoint. The consistency of the results across NLBS and
RSNA is encouraging, but it remains unknown whether the same improvement
would be obtained with other architectures, feature extractors, or
mammography pre-training strategies.

\item \textbf{Formal confidence bounds are limited to the case level.}
The Clopper--Pearson analysis assumes that each evaluated unit contributes
an independent outcome. Multiple images from the same examination are not
independent, so treating them as separate Bernoulli observations would not
support the same statistical interpretation. For this reason, CPU99 is
reported only for the case-level analysis. Image-level dismissal rates are
included as descriptive supporting results and should not be interpreted as
formally certified bounds.

\item \textbf{Clinical validation remains retrospective.}
The study evaluates model behavior retrospectively on existing datasets
and does not measure the effect of dismissal on radiologist workflow,
reading time, recall decisions, or patient outcomes. The reported dismissal
rates therefore establish technical potential under the defined operating
criteria, but prospective evaluation would be required before the method
could be used to support real screening decisions.

\end{itemize}
\section{Conclusion}

We introduced a closed-loop training approach for mammography dismissal
that repeatedly updates a provisional dismissal threshold during training
and uses this threshold to shape a positive-class penalty. The purpose is
to align model optimization more closely with the high-recall operating
region used for selective dismissal, while keeping final threshold
selection and statistical evaluation separate from the training process.

Across both NLBS and RSNA, the proposed model achieved the highest
case-level dismissal rate at the operating thresholds selected for the
98\% and 95\% recall targets and produced the lowest one-sided 99\%
Clopper--Pearson upper bound (CPU99) at each operating point. The
image-level results showed the same overall pattern in both in-domain
evaluations. The comparison with the fixed-$\tau$ ablation further
suggests that dynamically updating the dismissal boundary provides an
additional benefit beyond simply adding a dismissal-oriented loss term.

The results also showed that AUROC and dismissal performance do not
necessarily move together. The proposed model achieved the highest AUROC
on NLBS but the lowest among the evaluated configurations on RSNA, while
still achieving the highest case-level dismissal rates at both
recall-target operating points. This distinction is important because the
objective of the proposed method is not simply to improve overall ranking
performance, but to improve model behavior in the high-recall region
relevant to selective dismissal.

The same overall pattern was observed in the external
RSNA$\to$NLBS evaluation. The model weights were learned entirely from
RSNA and remained fixed. Operating thresholds were selected using the
20\% NLBS threshold-search subset and then applied without adjustment to
the independent 80\% evaluation subset. At the 98\% recall target, the
proposed model dismissed 12.95\% of NLBS cases with a CPU99 of 1.06\%,
whereas the CE-only baseline did not reach the required recall level. At
the 95\% recall target, the proposed model dismissed 19.87\% of cases
compared with 13.55\% for the baseline, while CPU99 was lower at 1.37\%
compared with 1.77\%.

Overall, these findings suggest that closed-loop threshold-aware training
can improve selective-dismissal performance relative to conventional
classification training and a fixed-threshold dismissal objective. The
reported statistical certificate applies specifically to the one-sided
upper confidence bound on cancer prevalence within the dismissed
evaluation cases and should not be interpreted as a guarantee of future
clinical safety. Further evaluation across additional screening
populations, model architectures, and prospective clinical workflows will
be needed to determine how reliably these findings translate to
real-world deployment.

\section*{Compliance with Ethical Standards}

This retrospective computational study used only publicly available,
de-identified mammography datasets. The Newfoundland and Labrador Breast
Screening (NLBS) dataset is publicly available through the Federated
Research Data Repository (FRDR), and the RSNA Screening Mammography Breast
Cancer Detection dataset is publicly available through the RSNA/Kaggle
challenge. No additional institutional ethical approval was required for
this secondary analysis of publicly available, de-identified data.

\section*{Conflict of Interest}

The authors declare no conflicts of interest.
\section*{Data and Code Availability}

The datasets used in this study are publicly available. The Newfoundland
and Labrador Breast Screening (NLBS) dataset is available through the
Federated Research Data Repository (FRDR)~\cite{Kendall2025NLBSdataset},
and the RSNA Screening Mammography Breast Cancer Detection dataset is
publicly available through the RSNA/Kaggle challenge~\cite{Trivedi2026-bz}.
No new clinical data were generated as part of this study.

\section*{Acknowledgments}

This work was supported by the Seed, Bridge, and Multidisciplinary Fund at
Memorial University of Newfoundland.

\bibliographystyle{unsrt}
\bibliography{cite}

\end{document}